\documentclass[runningheads]{llncs}
\usepackage[utf8]{inputenc} %
\usepackage[T1]{fontenc}    %
\usepackage{hyperref}       %
\usepackage{url}            %
\usepackage{booktabs}       %
\usepackage{amsfonts}       %
\usepackage{nicefrac}       %
\usepackage{microtype}      %
\usepackage{xcolor}  
\usepackage{color, colortbl}
\usepackage{fancyvrb}
\usepackage{epsfig}
\usepackage{graphicx}
\usepackage{wrapfig,lipsum,booktabs}
\usepackage{adjustbox}
\usepackage{mathtools}
\usepackage{caption}
\usepackage{xspace}
\usepackage{algorithm}
\usepackage{algpseudocode}
\usepackage{listings}
\usepackage{import}
\usepackage{subcaption}
\usepackage{multirow}

\DeclareCaptionSubType*{algorithm}

\DeclareCaptionLabelFormat{alglabel}{Alg.~#2}

\newcommand{\JC}[1]{{\color{blue}{\small\bf\sf [John: #1]}}}

\definecolor{Gray}{gray}{0.9}

\def\cryoRL{cryoRL\xspace}

\def\zapcolorreset{\let\reset@color\relax\ignorespaces}
\def\colorrows#1{\noalign{\aftergroup\zapcolorreset#1}\ignorespaces}

\newcommand{\mycomment}[1]{}

\begin{document}
\title{CryoRL: Reinforcement Learning Enables Efficient Cryo-EM Data Collection} %

\titlerunning{CryoRL: Reinforcement Learning Enables Efficient Cryo-EM Data Collection}
\author{Quanfu Fan\inst{1,6} 
\and
Yilai Li\inst{2,6} 
\and
Yuguang Yao\inst{3}
\and
John Cohn\inst{1}
\and 
Sijia Liu\inst{3}
\and 
Seychelle M. Vos\inst{4,7}
\and 
Michael A. Cianfrocco\inst{2,5,7}
}

\authorrunning{Q. Fan et al.}
\institute{
MIT-IBM Watson AI Lab, Cambridge, MA USA 
\and
Life Sciences Institute, University of Michigan, Ann Arbor, MI USA
\and
Department of Computer Science and Engineering, Michigan State University, East Lansing, MI USA
\and
Department of Biology, Massachusetts Institute of Technology, Cambridge, MA USA 
\and 
Department of Biological Chemistry, Michigan Medicine, University of Michigan, Ann Arbor, MI USA
\and 
Equal contributions
\and 
For correspondence: M.A.C. \email{mcianfro@umich.edu} \& S.M.V. \email{seyvos@mit.edu}
}

\maketitle

\begin{abstract}
Single-particle cryo-electron microscopy (cryo-EM) has become one of the mainstream structural biology techniques because of its ability to determine high-resolution structures of dynamic bio-molecules. However, cryo-EM data acquisition remains expensive and labor-intensive, requiring substantial expertise. Structural biologists need a more efficient and objective method to collect the best data in a limited time frame. We formulate the cryo-EM data collection task as an optimization problem in this work. The goal is to maximize the total number of good images taken within a specified period. We show that reinforcement learning offers an effective way to plan cryo-EM data collection, successfully navigating heterogenous cryo-EM grids. The approach we developed, cryoRL, demonstrates better performance than average users for data collection under similar settings. 
\end{abstract}

\section{Introduction}
\label{sec:intr}
Single-particle cryo-electron microscopy (cryo-EM) has become one of the mainstream structural biology techniques due to its ability to solve the structures of many bio-molecules with moderate heterogeneity and without the need for crystallization. In recent years, continued software development has led to automation in both data collection and image processing \cite{baldwin2018big}. Moreover, with the improvement of the detectors and microscopes techniques, data acquisition has been dramatically accelerated \cite{cheng2018high,weis2020combining}. 

\begin{wrapfigure} {lr}{0.5\textwidth}
    \centering
    \includegraphics[width=0.8\linewidth]{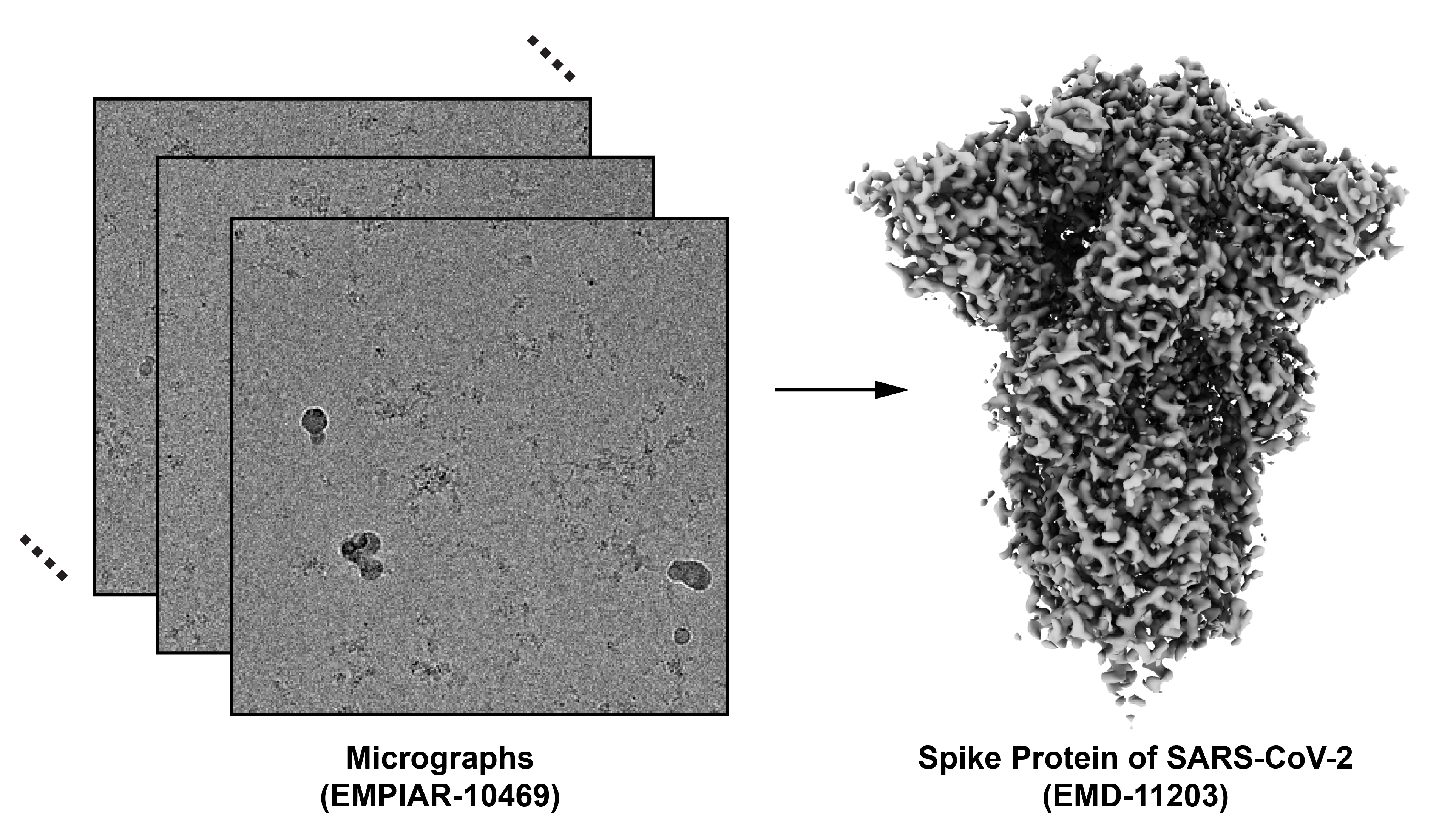}
    \caption{\small{Cryo-EM structure of the SARS-CoV-2 spike protein.}}
    \label{fig:figure1} 
\end{wrapfigure}

Cryo-EM serves as a critical tool in the development of vaccines and therapeutics to combat COVID-19 by SARS-CoV-2 (Fig.~\ref{fig:figure1}). Within weeks of the release of the genomic sequence of SARS-CoV-2, cryo-EM determined the first SARS-CoV-2 spike protein structure \cite{wrapp2020cryo}. Since this original publication, cryo-EM was used to determine additional SARS-CoV-2 structures such as spike protein bound to antibody fragments \cite{lempp2021lectins,scheid2021b}, remdesivir bound to SARS-CoV-2 RNA-dependent RNA polymerase \cite{bravo2021remdesivir,yin2020structural,kokic2021mechanism}, and reconstructions of intact SARS-CoV-2 virions \cite{yao2020molecular,ke2020structures}.

\setlength\intextsep{0pt}

Despite these advances, cryo-EM data collection remains \emph{ad-hoc}, rudimentary, and subjective. Due to variations in sample quality across a cryo-EM grid, users collect images at different magnifications ranging from resolutions of 0.66 mm to 500 Å.  Significant user expertise enables experts to define and refine locations suitable for data collection. To provide objective feedback,  "on-the-fly" image processing \cite{lander2009appion, tegunov2019real} can confirm high-quality regions on the cryo-EM sample. Despite this information, data collection remains highly subjective. Cryo-EM is also an expensive technique, further compounding challenges faced by users. Purchasing, preparing, and installing a top cryo-electron microscope can cost about \$10 to 20 million USD, and the daily operational cost can be around \$10,000 USD \cite{Hand2020WeNA}. Therefore, structural biologists need methods that can help collect the best data possible in a limited amount of time.

In this paper, we formulate the data collection problem as an optimization task where the goal is to learn intelligent strategies from data to guide the microscope movement, possibly via manual suggestion or robotic manipulation.
 We model the optimization problem as a Markov decision process and propose to solve it by combing supervised classification and deep reinforcement learning (RL)~\cite{Sutton1998}. We present a new data acquisition algorithm, cryoRL, which enables data collection with no subjective decisions,  no user intervention, and increased efficiency. To address the potential enormously large action space in our problem, we further propose to eliminate irrelevant or sub-optimal actions based on the classification results to enable effective policy exploration, which improves cryoRL in both efficiency and accuracy.
As compared with human subjects, cryoRL achieves better performance than average users. To the best of our knowledge, cryoRL is the first AI-based algorithm in cryo-EM data acquisition such that a policy is learned and can directly help the user steer the microscope.  
\mycomment{
To design, implement, and test cryoRL, we collected an "unbiased" dataset. The first of its kind, our unbiased dataset includes 4017 micrographs from 31 squares on a single grid.\JC{jc - i think we need to breifly describe the greid, square, hole hierarchy before saying this.. no ? }\ Data collection involved no user decisions; instead, we selected the areas from a systematic pattern of data collection to obtain images of all holes and micrographs in 31 squares.  Our unbiased dataset will be released to the public to serve as a critical benchmark for the evaluation of cryo-EM data collection algorithms.
}

We collected datasets on different grid types to design, implement, and test cryoRL. The first of its kind, our data collection involves no user decision; instead, we selected the areas from a systematic pattern of data collection to obtain images of all holes and micrographs~(See Section ~\ref{sec:data-collection} for details of our data).  Our datasets will be released to the public to serve as a critical benchmark for evaluating cryo-EM data collection algorithms.

To summarize, our high level conclusions and contributions include: 
\begin{itemize}
  \item CryoRL achieves better performance than other popular optimization techniques such as Genetic Algorithm~\cite{weise2009global} and Simulated Annealing~\cite{kirkpatrick1984optimization}, and demonstrates good generalization capability;
    \item CryoRL with our proposed invalid action elimination runs 2$\sim$3 times faster than the vanilla DQN baseline %
    while enabling more robust policy learning;
    \item CryoRL offers a new approach to cryo-EM data collection that demonstrates promising results by outperforming average users in a human performance study; %
  \item We are providing a first-kind-of cryo-EM dataset that is critical for algorithm development and benchmarking. %
    
\end{itemize}

\section{Related Work}
\label{sec:literature}
There are currently no automated, 'intelligent' cryo-EM data collection approaches. Instead, subjective decision-making drives cryo-EM data acquisition. To guide user-driven data collection, on-the-fly image analysis provides feedback on data quality, including Appion \cite{lander2009appion}, Warp\cite{tegunov2019real}, and cryoSPARC Live. To provide more objective measures of data quality to users, researchers have developed a pretrained deep learning-based micrograph assessment model \cite{li2020high} and downstream on-the-fly data processing \cite{stabrin2020transphire}. However, despite these efforts, on-the-fly processing requires a sizeable number of micrographs before providing useful feedback. Data collection requires user training to develop expertise to guide data collection in the most efficient manner possible.

Reinforcement learning (RL) has been widely applied to address practical optimization problems such as network planning~\cite{SIGCOMM-HZhu}, vehicle routing~\cite{nazari2018reinforcement}, on-line recommendation~\cite{KDD-2018-ZhaoZDXTY}, robot trajectory optimization~\cite{doi:10.1177/0278364907087426} as well as game playing~\cite{SilverHuangEtAl16nature}. Some practical applications have adopted RL for enabling fast data collection or processing. For example in \cite{tong2020deep,zhang2020hierarchical}, RL-based methods are proposed to optimize unmanned aerial vehicle flight trajectories for efficient data collection. In~\cite{krull2020artificial}, a deep RL agent is used to condition the state of the probe for autonomous Scanning Probe Microscopy (SPM). Nevertheless, automating data collection by machine learning techniques in real-world scenarios remains an under-explored problem.

Large action spaces are a common problem to deal with in RL. Existing techniques include action masking~\cite{berner2019dota,ye2020mastering} to mask out invalid actions, action elimination~\cite{zahavy2018learn} to remove inferior actions, and action reshaping~\cite{kanervisto2020action} to transform a discrete action space to a simpler one or a continuous one. Our proposed action elimination is in a similar spirit to Action Elimination Network (AEN)~\cite{zahavy2018learn}, but instead relies on the estimated quality of a hole to exclude invalid actions directly rather than learning to reduce the action space. 
\section{Cryo-EM Data Collection}
\label{sec:data-collection}
\begin{figure}[!ptb]
    \centering
    \includegraphics[width=0.75\linewidth]{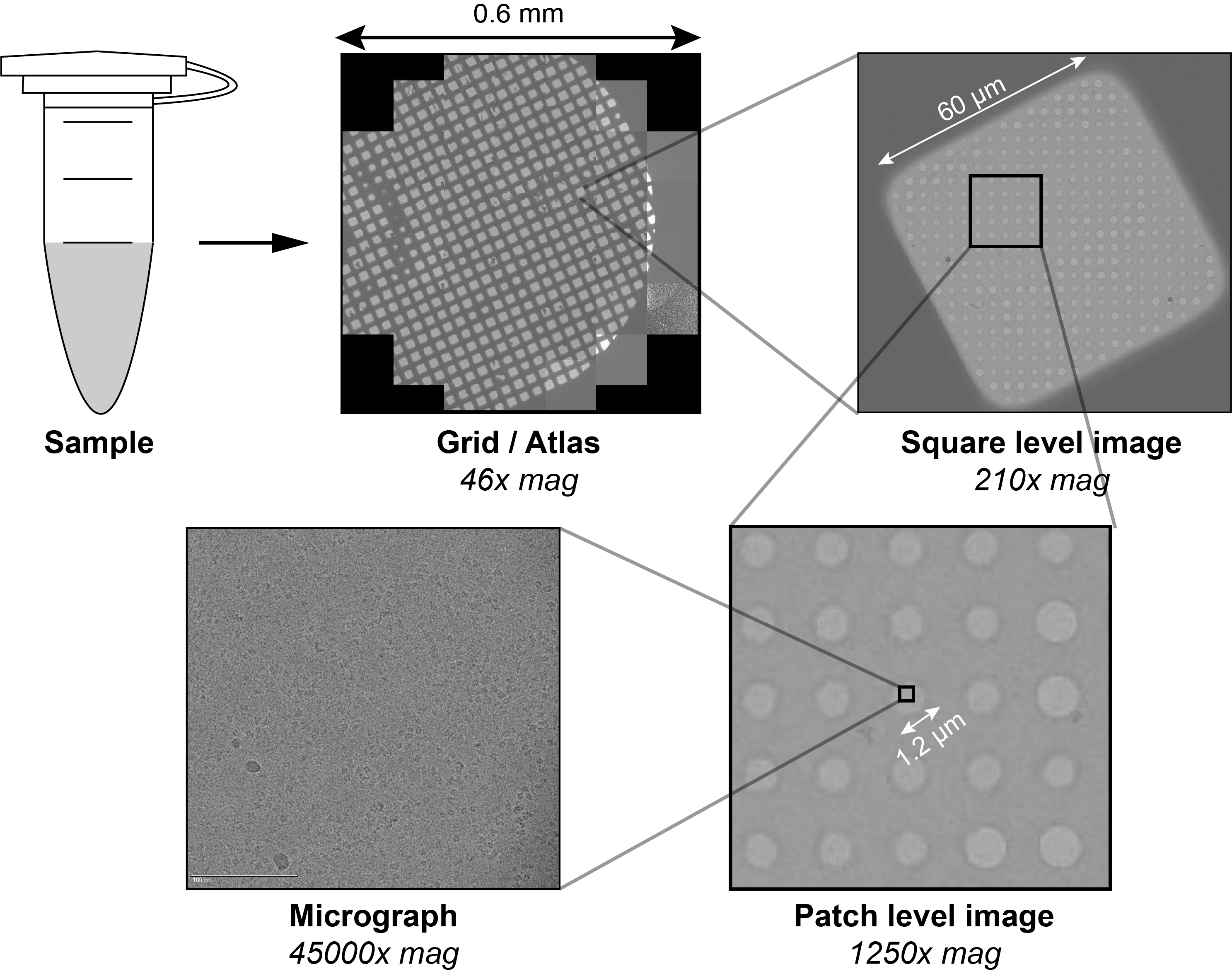}
    \caption{\small{Overview of cryo-EM data collection. A purified sample is prepared and vitrified on the support grid. The atlas image provides a low magnification overview by stitching multiple "grid-level" images into a single montage. Next, users will select specific squares to image at medium magnification. After inspection, the user selects "patch" areas on the square to inspect holes with higher magnification, using the patch image to decide holes to collect for micrographs. The micrographs contain high-resolution images for downstream data processing. %
    }}
    \label{fig:figure2} 
\end{figure}

The general practice of data acquisition in cryo-EM is abstracted in Fig.~\ref{fig:figure2}. Typically, a purified biological sample is dispensed and vitrified onto a grid comprised of gold or copper support bars. A grid contains a mesh of squares, and each square has a lattice of regularly-spaced holes. Ideally, within each hole, there are vitrified single-particles related to the sample of interest, where data collection amounts to users recording images of each hole as micrographs. 

Cryo-EM samples exhibit heterogeneity across the specimen. Whereas there are many local correlations between squares and holes on the grid, many holes are empty, contain aggregates, or contain non-vitreous ice contamination. The user has no prior knowledge of such distribution until the square-level or hole-level images are acquired, which can be captured by the microscope by changing to different magnifications. Note that each greater magnification requires significant time for the microscope to move and settle. Moreover, because the time on the microscope is precious and limited, data collection can typically cover less than 1$\%$ of the total grid. The user needs to navigate through the "grid-square-hole" hierarchy and collect the best micrographs in a limited time.

\setlength\intextsep{0pt}

In this paper, we suppose that a user preselects a set of squares and patch-level images by a quick atlas survey. We formulate the data acquisition task to find the highest quality holes and plan the overall data collection route with cryoRL. Although this is not the traditional way people collect cryo-EM data, we believe such prerequisites provide a global understanding of the hole quality distribution on a grid by taking a series of low to medium magnification square and patch level images. 

Each micrograph has an objective measure of data quality, which is the goodness-of-fit for the frequency domain when estimating the defocus of the micrograph. We introduce the term "CTFMaxRes" to be the maximum resolution (Å) for the fit of the contrast transfer function (CTF) to a given micrograph using the program CTTFIND4 \cite{rohou2015ctffind4}.  CTFMaxRes is calculated from the 1D power spectrum of the micrograph and estimates the maximum resolution for the detected CTF oscillations \cite{brahme2014comprehensive}. The field of cryo-EM utilizes CTFMaxRes to provide an indirect metric for data quality. In general, the lower this value, the higher the quality of the micrograph. CryoRL will predict the quality of each hole from the patch-level image using an image classifier (Section~\ref{sec:approach}). For simplicity, we define CTFMaxRes as the CTF value for this paper.

\section{RL-based Approach}
\label{sec:approach}

\subsection{Overview}
\begin{figure}[!ptb]
    \centering
    \includegraphics[width=1\linewidth]{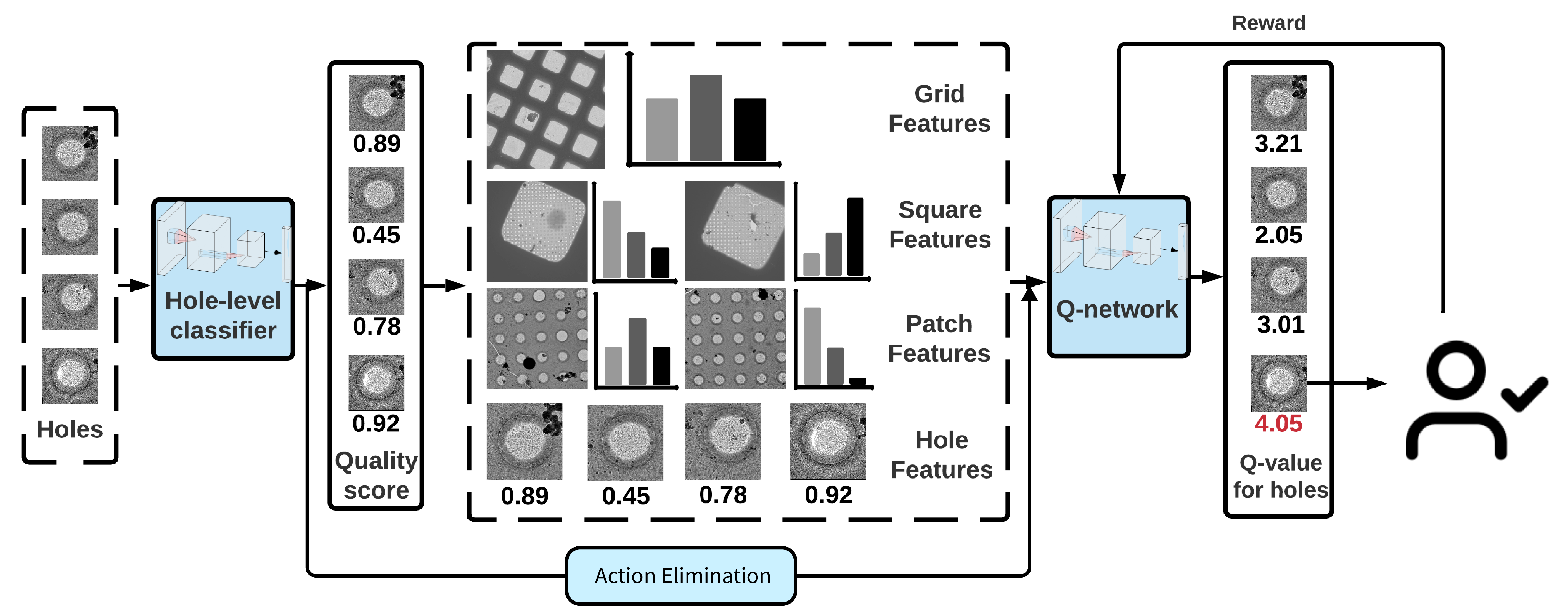}
    \caption{\small{cryoRL-guided cryo-EM data collection. cryoRL consists of an offline hole-level classifier to estimate the quality of holes and a deep Q-network to learn effective strategies for steering the microscope for data collection. The hole classifier outputs serve as features of the RL network. The agent (or user) provides feedback (rewards) to the system according to the microscope movement and the measured quality of the micrographs taken for the holes recommended by the system. In addition, to address the potential issue of large action spaces in our problem, we developed an efficient method to eliminate invalid actions based on the quality of holes.}}
    \label{fig:RL_system} 
    \vspace{-3mm}
\end{figure}

During data collection, a user needs to make decisions based on the quality of images taken at different magnification levels: grid, square, and patch-level. Given that the data are visually similar and there are significant costs (time) of moving to other grid areas and refocusing, there is no easy planning that a user can make manually in a regular data collection. As a result, the user explores only a small portion of a grid, making the data collection process inefficient and subjective. 

In this work, we formulate the data collection problem as planning an optimal path for operating the microscope. The goal is to move the microscope to explore desired places on a grid in a given amount of time, with the operational cost constraints taken into account~(Section~\ref{subsec:problem}). We propose to solve the path planning problem by RL, a technique that has demonstrated success in many vision applications~\cite{DBLP:journals/corr/abs-2108-11510}. Compared to other widely used optimization solvers such as Genetic Algorithm (GA)~\cite{weise2009global} and Simulated Annealing (SA)~\cite{kirkpatrick1984optimization}, RL is more suitable for modeling sequential problems and possibly less heuristic in system design.

As illustrated in Fig.~\ref{fig:RL_system}, our proposed approach combines an image classifier and an RL network to enable automatic planning of microscope movement. The supervised classifier categorizes a hole into low or high quality based on its CTF value. Efficient hierarchical feature representations for cryo-EM images at different magnification levels are generated from the classification results. These features, along with the observation history, are exploited to train a deep Q-network (DQN)~\cite{mnih2013playing} to assess the status of all the unvisited holes and suggest the best holes to look at next. We further design a rewarding mechanism to drive the learning of DQN. The design in general values small microscope movements to avoid wasted time. For example, moving to a different patch on the same grid-level image receives a higher score than changing to an entirely new grid-level image~(Section~\ref{subsec:dqn}). Finally, to handle the potentially large action space in our problem, we propose a method to eliminate invalid actions, which not only results in a significant speedup of CryoRL by 2$\sim$3 times, but also improves the robustness of the approach (Section~\ref{subsec:dqn}).

As mentioned earlier, a human user can usually cover a small portion of the grid during a data collection session. In contrast, one significant advantage of our proposed approach is that it allows for a substantially larger exploration of the grid by the microscope as the approach learns to focus on promising regions with high-quality data. We demonstrate in Section~\ref{sec:expr} that our system is highly effective, achieving comparable performance to human subjects.

\subsection{Problem Formulation}
\label{subsec:problem}
\setlength\intextsep{0pt}
\begin{wrapfigure}{lr}{0.5\textwidth}
    \centering
    \includegraphics[width=0.75\linewidth]{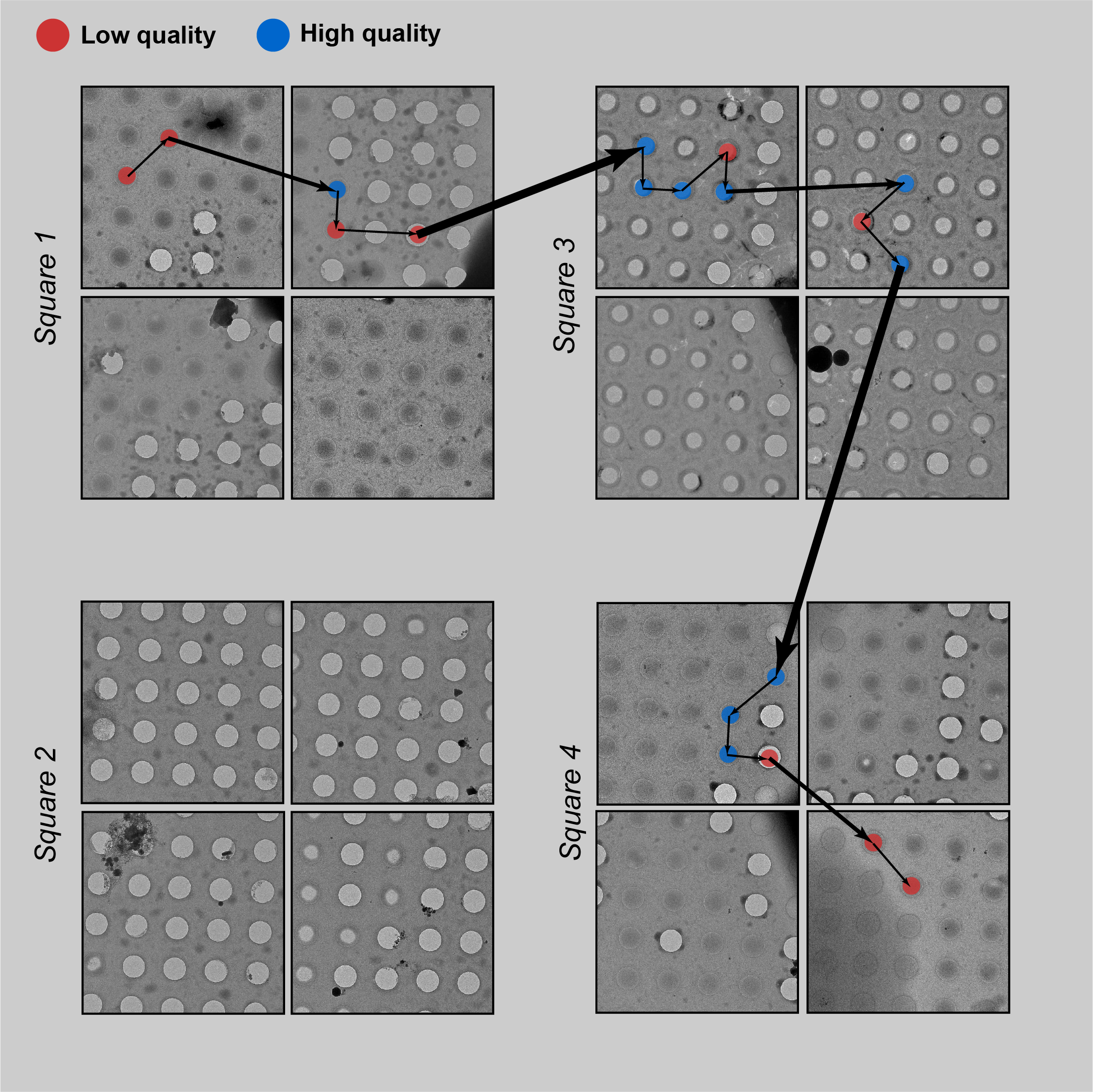}
    \caption{\small{A schematic illustration of a path showing the microscope movement planned in data collection. Different microscopic operations are associated with different costs, which are indicated by the edge width. }}
    \label{fig:RL_path} 
\end{wrapfigure}

As previously described, cryo-EM data collection is steering the microscope hierarchically at different magnification levels to explore a grid to identify high-quality micrographs. This sequential process involves several mechanical operations to allow microscope navigation to different regions of a grid. The process of data collection involves area switching (changing to a new grid-level image), square switching (changing to a new square-level image), and patch switching (changing to a new patch-level image). Since the data distribution is non-uniform on a grid and it takes time to prepare the microscope for imaging at different levels, an automatic method to guide the data exploration more intelligently will improve data quality and efficiency for data collection.

As shown in Fig.~\ref{fig:RL_path}, an effective data collection session aims at finding a sequence of holes where there is a considerable portion of high-quality micrographs.
Let $\mathcal{H}=\{h_l|l=1\cdots n_h\}$ be a sequence of holes in a set of patches $\mathcal{P}$ sampled from different square-level and grid-level images ($\mathcal{S}$ and $\mathcal{G}$) by the user. We denote $\mathcal{P}_{h_l}$, $\mathcal{S}_{h_l}$ and $\mathcal{G}_{h_l}$ as the corresponding patch-level, square-level and grid-level images of $h_l$, respectively. Also, $ctf(h_l)$ is a function representing the CTF value of a hole $h_l$. Our goal is to identify a maximum subset of holes from $\mathcal{H}$ with low-CTF values in a given amount of time $\tau$. 
Mathematically, this is equivalent to optimizing an object function as follows, \\
\begin{equation}
\max \sum_{l=0}^{n_h-1} {(\rho(h_l) - c(t(h_l)))} \quad \textrm{s.t. } \sum_{l=0}^{n_h-1} {t(h_l) \le \tau}
\label{eq:objective}
\end{equation}
where $\rho(h_l)$ be such an indicator function for a hole $h$ that 
\begin{equation}
  \rho(h_l) =
    \begin{cases}
      1 & \text{if $ctf(h_l) \le 6.0$}\\
      0 & \text{otherwise}
    \end{cases}       
\end{equation}
and $c$ is a cost associated with the corresponding microscope operation and determined by the total amount of time $t(h_l)$ spent on $h_l$.  In this work, we define $t(h_l)$ in minutes by the movement of the microscope, i.e,
\[
\small
t(h_l) =
    \begin{cases}
      2.0 & \text{if $\mathcal{P}_{h_{l-1}}= \mathcal{P}_{h_l}$ (same patch)}\\
      3.0 & \text{if $\mathcal{P}_{h_{l-1}} \ne \mathcal{P}_{h_l}, \mathcal{S}_{h_{l-1}}= \mathcal{S}_{h_l}$ (same square)}\\
      5.0 & \text{if $\mathcal{S}_{h_{l-1}} \ne \mathcal{S}_{h_l},  \mathcal{G}_{h_{l-1}}=\mathcal{G}_{h_l}$ (same grid)}\\
      10.0 & \text{if $\mathcal{G}_{h_{l-1}} \ne \mathcal{G}_{h_l}$ (different grid)}\\
    \end{cases}       
\]
Note that the time $t$ above is set in a way so that it highly corresponds to the natural time of the microscope movement in real-world scenarios. Nevertheless, in practice, it can be more precisely calculated based on the distance of the microscope movement and other factors.

By setting $r(h_l)=\rho(h_l)-c(t(h_l)))$, we can further rewrite Eq.~\ref{eq:objective} as 
\begin{equation}
\max \sum_{l=0}^{n_h-1} {r(h_l)} \quad \textrm{s.t. } \sum_{l=0}^{n_h-1} {t(h_l) \le \tau}
\label{eq:rl-objective}
\end{equation}

Eq.~\ref{eq:rl-objective} has the same form as the standard accumulative reward (without a discount factor) that is maximized in RL ~\cite{Sutton1998}. In what follows, we describe how to design a RL system to solve the path optimization problem in Eq.~\ref{eq:rl-objective}.

\subsection{Path Optimization by Reinforcement Learning}
\label{subsec:dqn}
We study the cryo-EM data acquisition task by RL, where an agent interacts with environment (i.e. the grid here) by sequentially selecting holes for taking micrographs over a sequence of time steps, with an objective to maximize the cumulative reward described in Eq.~\ref{eq:rl-objective}. We briefly describe the basic components of our system as follows.

    \indent \textit{Environment}: the atlas or grid.\\
    \indent \textit{Agent}: a robot or user steering the microscope.\\
  \indent \textit{States}. Let $u_i \in \{0, 1\}$ be a binary variable denoting the status of hole, i.e. visited or unvisited. Then a state $s$ in our setting can be represented by a sequence of holes and their corresponding statuses $s=<(h_1, u_1), (h_2, u_2), ..., (h_{n_h},u_{n})>$ where $n$ is the total number of holes. \\
  \indent \textit{Actions.} An action $a_i$ of the agent in our system is to move the microscope to the next target hole $h_i$ for imaging. Note that in our case, any unvisited hole has a chance to be picked by the agent as a target, thus the action space is large. Also, during tests, the number of holes (i.e actions) is unknown. Instead of adopting more sophisticated methods to handle continuous action space as proposed in~\cite{NIPS2007_0f840be9,pmlr-v80-lee18b}, we simply modify the Q-network to estimate the Q-value for every single hole rather than all of them at once. We show this suffices for handling the large action space in our case.   \\
  \indent \textit{Rewards.} We assign a positive reward 1.0 to the agent if an action results in a target hole with a CTF value less than 6.0Å and 0.0 otherwise. The agent also receives a negative reward depending on the operational cost associated with a hole visit. Specifically, we model the negative reward as $c(h_l) =1.0-e^{-\beta (t(h_l)-t_0)} (\beta > 0, t_0\ge 0)$. We empirically set $\beta$ and $t_0$ to 0.185 and 2.0, which define the final reward function for our RL system as,
  \[
\small
  r(a_i) =
    \begin{cases}
      1.0 & \text{if $ctf(h_l)<6.0$ \& $\mathcal{P}_{h_{i-1}}= \mathcal{P}_{h_i}$ }\\
      0.57 & \text{if $ctf(h_l)<6.0$ \& $\mathcal{P}_{h_{i-1}} \ne \mathcal{P}_{h_i}$ \& $\mathcal{S}_{h_{i-1}}= \mathcal{S}_{h_i}$} \\
      0.23 & \text{if $ctf(h_l)<6.0$ \&  $\mathcal{S}_{h_{i-1}} \ne \mathcal{S}_{h_i}$ \& $ \mathcal{G}_{h_{i-1}}=\mathcal{G}_{h_i}$ }\\
      0.09 & \text{if $ctf(h_l)<6.0$ \& $\mathcal{G}_{h_{i-1}} \ne \mathcal{G}_{h_{i}}$}\\
     0.0 & otherwise \\   
     \end{cases}       
\]
Note that the design principle of these rewards is to reward more small microscope movement. As shown later in the experiments (Section~\ref{subsec:ablation}), CryoRL is not sensitive to the changes of the rewards as long as the design described above is followed.

\mycomment{
\textbf{Environment}: the atlas \\
\textbf{Agent}: \\
\textbf{States}. Let $a_i \in \{0, 1\}$ be a binary variable to indicate whether or not a hole is visited. Then a state $s$ in our setting can be represented by $s=<a_1, a_2, ..., a_{n_h}>$. \\
\textbf{Actions.} \\
\textbf{Rewards.} 1 if the ctf of a hole is less than 4.0 or 0 otherwise. There is also an operational cost associated with each visit, i.e. a negative reward.
question: instead of using instant reward for each visited hole with low CTFs, can we use an accumulated reward for a sequence of holes visited within a pre-specified time period?
}

\noindent \textbf{Deep Q-learning} We apply the deep Q-learning approach proposed in \cite{mnih2013playing} to learn our policy for cryo-EM data collection.
The goal of the agent is to select a sequence of actions (i.e. holes) based on a policy to maximize future rewards (i.e the total number of low-CTF holes). In Q-learning, this is achieved by maximizing the action-value function $Q^{\ast}(s,a)$, i.e. the maximum expected return achievable by any strategy (or policy) $\pi$ , given an observation (or state) $s$ and some action $a$ to take. In other words, $Q^{\ast}(s,a) = \max_{\pi} E [R_t|s_t = s,a_t = a,\pi]$ where $R_t=\sum_t^{\infty}\gamma^{t-1} r_t$ is the accumulated future rewards with a discount factor $\gamma$. Q* can be found by solving the Bellman Equation~\cite{Sutton1998} as follows, 
\begin{equation}
Q^{*}(s,a)=E_{s'}[r+\gamma\max_{a'}Q^{*}(s',a')|s,a]
\label{eq:Q-equation}
\end{equation}
In practice, the state-action space can be enormous,
thus in~\cite{mnih2013playing}, a deep neural network parameterized by $\theta$ is applied to approximate the action-value function. The network is referred to as Deep Q-Network (DQN) in the original paper. DQN can be trained by minimizing the following loss functions $L(\theta)$,
\begin{equation}
L(\theta)=E_{s,a,r,s'}[(y-Q(s,a;\theta)^2]
\label{eq:L-equation}
\end{equation}
where $y=E_{s'}[r+\gamma\max_{a'}Q(s',a')|s,a]$ is the target for the current iteration.
The derivatives
of the loss function $L(\theta)$ are expressed as follows:
\begin{equation}
\small
\nabla _{\theta} L(\theta)=E_{s,a,r,s'}[(r+\gamma \max_{a'} Q(s',a'; \theta') - Q(s,a;\theta)) \nabla _{\theta}Q(s,a;\theta)]
\end{equation}
Experience replay is further adopted in ~\cite{mnih2013playing} to store into memory the transition  at each time-step, i.e $(s_t,a_t,r_t,s_{t+1})$, and then sample the stored samples for model update during training.

\begin{minipage}[!bt]{\textwidth}
\begin{minipage}[b]{0.48\textwidth}
    \centering
    \includegraphics[width=0.75\linewidth]{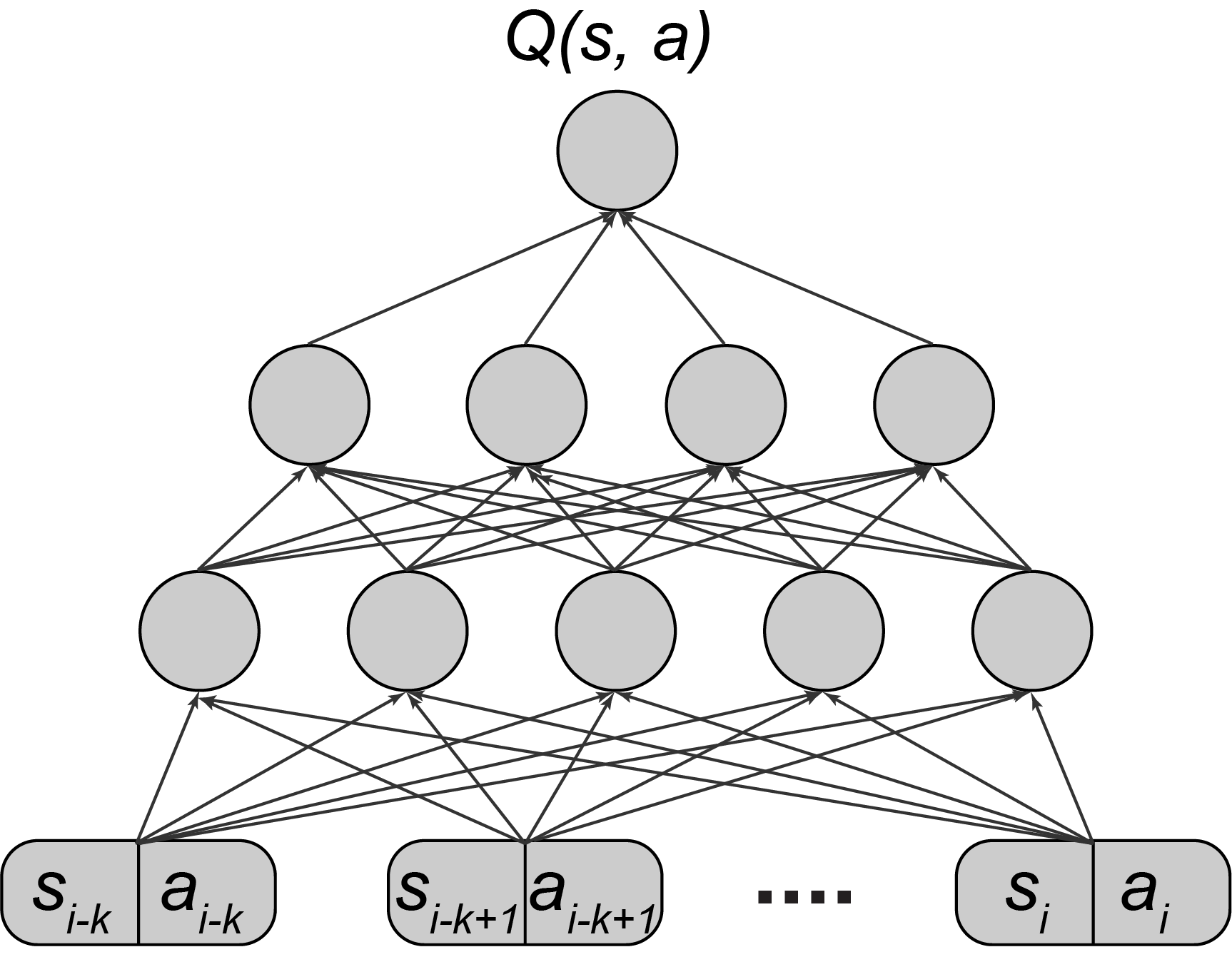}
    \captionof{figure}{\small{The architecture of DQN. The network has only one single output node to estimate the Q-value for an action-state pair. }} 
    \label{fig:DQN} 
    \end{minipage}
    \hfill
      \begin{minipage}[b]{0.48\textwidth}
    \centering
    \includegraphics[width=0.6\linewidth]{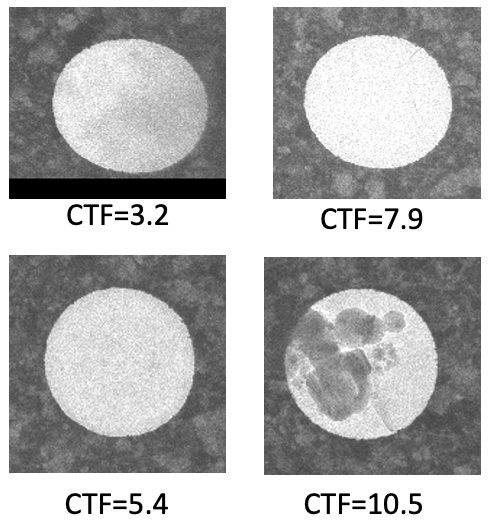}
    \captionof{figure}{\small{Examples of hole images and their CTF values. A hole with a CTF $\le$ 6 is considered good in our paper.}}
    \label{fig:hole_example} 
  \end{minipage}
  
\end{minipage}

\mycomment{
\begin{wrapfigure}{lr}{0.5\textwidth}
    \centering
    \includegraphics[width=0.75\linewidth]{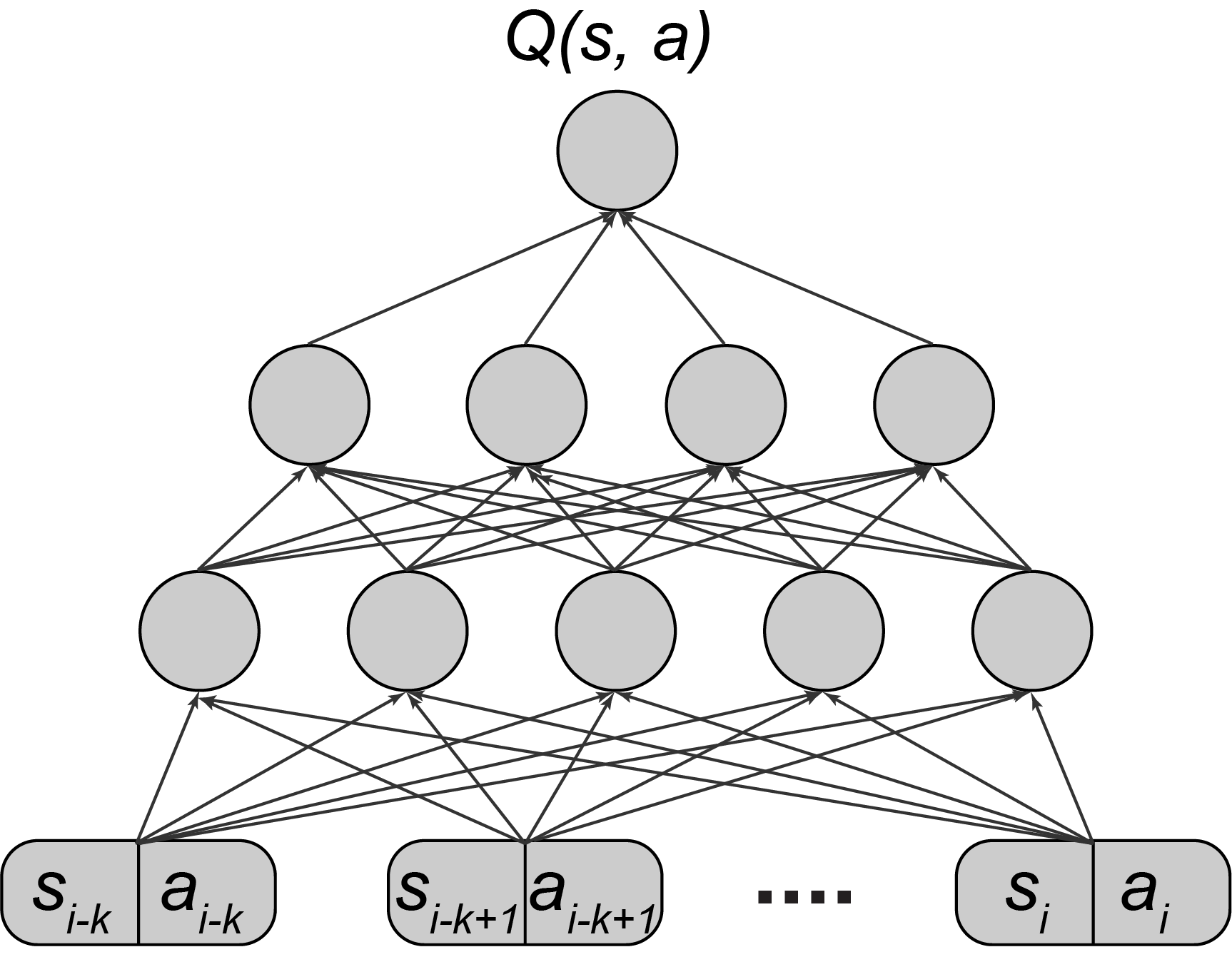}
    \caption{\small{The architecture of DQN. The network has only one single output node to estimate the Q-value for an action-state pair. }}
    \label{fig:DQN} 
\end{wrapfigure}
}

\textbf{DQN with Action Elimination via Patch Ranking} 
In a regular scenario such as playing Atari \cite{mnih2013playing} where the action space is small and fixed, a network can be trained to predict all the actions at once. However, this is not suitable for our case as our action space is not fixed and can grow large depending on the training data size. To deal with this issue, we modify the Q-network to predict the Q-value for each hole (i.e., action) using one single output, as shown in Fig.~\ref{fig:DQN}. The Q-value for all the actions can then be batch processed and the $\epsilon$-greedy scheme is applied for action selection. The DQN used in our work is a 3-layer fully connected network. The size of each layer is 128, 256 and 128, respectively.

 The potentially enormously large action space in our problem makes policy exploration quite inefficient as most actions sampled from such a space are not useful. To avoid executing too many sub-optimal actions in learning, we propose an effective method to reduce the action space by restricting the valid actions to a small portion of holes predicted by the classifier as low-CTFs. We start by ranking all the grid-level images by their numbers of low-CTF predictions, from high to low, and then the patches in the same way. The high-ranked patches are likely to contain more valid holes and should be visited more frequently during the learning. The pre-specified duration (i.e. $\tau$ in Eq.~\ref{eq:objective}) as well as the switching costs $t(h_l)$ defined in Section~\ref{sec:approach} allow us to obtain an upper limit $N_{max}$ of good holes if \textit{all the holes are assumed to be low CTF} and visited in the sorted order described above. We then select a minimum set of patches $P$ with a total number of low-CTF predictions $\ge$ $\beta N_{max}$ ($\beta > 0$), and all the holes in $P$ define a reduced new space for Q-learning. Here, $\beta$ is a user-defined parameter to control the size of the valid action set. Our approach is not sensitive to $\beta$, and any number between 1.0$~\sim$2.5 works reasonably well. We thus empirically set $\beta$ to $1.5$ in tests and a larger number in training to enlarge the exploration space for CryoRL. 

 The details of our algorithm can be found in the appendix. %
 Unlike the Elimination Network (EAN) proposed in~\cite{zahavy2018learn}, our approach redefines the action space before Q-learning, so it can be applied to any policy learners without modification of them. We show later in the experiments that our approach results in a significant speedup of 2$\sim$3 times over the vanilla DQN and improves other policy learners such as A2C~\cite{a2c} and C51~\cite{bellemare2017distributional} remarkably.

\mycomment{
\begin{algorithm}
\caption{Fast CryoRL with Action Elimination}\label{alg:fast-CryoRL}
\begin{algorithmic}
\Require States $S$, Actions $A$, Rewards $R$,\\ Learning Rate $\alpha$, Discounting factor $\gamma$, Switching costs $C$, Duration $\tau$
\Procedure $QLearning\_AE(S, A, R, \alpha, \gamma)$

\State $P \gets [p_0, p_1, \cdots, p_n]$ \Comment{Patches}
\State $L \gets [l_0, l_1, \cdots, l_n]$ \Comment{\# of predicted lCTFs in each patch}
\State $A' \gets ActionElimination(P,L,C, \tau)$
\State $Q \gets QLearning(S, A', R)$

\Return $Q$

\Procedure $ActionElimination(P, L, C, \tau)$
\State $N_{max} \gets Max\_lCTF($P$, $C$, \tau)$ \Comment{maximum lCTFs found in a perfect case}
\State $n \gets 0$
\State $A' \gets \{\}$
\For{$p_i$ in $P$}
    \State $n \gets n + l_i$
    \State $A' \gets A'\bigcup \{h_j\in p_i| j=1 \cdots m_i\}$
    \If {$n \ge N_{max}$}
        \State $break$
    \EndIf
\EndFor

\Return $A'$

\end{algorithmic}
\end{algorithm}
}

\setlength\intextsep{0pt}
\begin{table}[t]
    \centering
    \small
    \begin{adjustbox}{max width=0.9\linewidth}
    \begin{tabular}{c|c|c}
        \toprule
Feature Type & Definition & Value \\
\midrule
hole & is it low-CTF?\\
           & is it visited? & \{0,1\}\\
\midrule
           & \# of unvisited holes &\\
patch/square/grid       & \# of unvisited lCTFs & 0$\sim150^{*}$\\
         & \# of visited holes & \\
       & \# of visited lCTFs & \\
\midrule
 & a new patch-level image? & \{0,1\}\\
microscope  & a new square-level image? & \{0,1\}\\
 movement& a new grid-level image? & \{0,1\}\\
    \bottomrule
      \multicolumn{3}{l}{\footnotesize $^*$: the maximum number of holes allowed in a grid-level image in our setting} 
    \end{tabular}
    \end{adjustbox}
    \vspace{-1mm}
    \caption{\small Input features to DQN}
    \label{table:dqn-feaure} 
\end{table}

\textbf{Features to DQN} The quality of a hole is directly determined by its CTF value. Similarly, the number of low-CTF holes (lCTFs) in a hole-level image indicates the quality (or value) of the image, and a good RL policy should always consider prioritizing high-quality patches first in planning. The same holds true for square-level and grid-level images. Based on this, we design hierarchical input features to the DQN according to the quality of images at different levels. We also consider the information of microscope movement as it tells whether the microscope is exploring a new region or staying at the same region. The details of these features can be found in Table~\ref{table:dqn-feaure}. Finally, a sequence of these features for the last $k-1$ visited holes as well as the current one to be visited are concatenated together to form the input to DQN. In our experiments, $k$ is empirically set to $4$.

\noindent\textbf{Hole-level Classification} We trained the hole-level classifier offline by cropping out the holes in our data using the location provided in the meta data. 
Fig.~\ref{fig:hole_example} illustrates a few examples of hole images. These images are actually visually ambiguous, confounding the task of building generalized hole classifiers, as shown in Section~\ref{subsec:main_results}.
Using an offline classifier enables fast learning of the Q function as only the Q-network is updated in training and its input features can be computed efficiently. However, it is possible to jointly learn the classifier and DQN to further improve performance. We leave this possibility for future work.

\section{Experiments}
\label{sec:expr}
\subsection{Experimental Setup}
\noindent\textbf{Dataset}
To design and evaluate the performance of cryoRL, we collected an "unbiased" cryo-EM dataset (\textbf{Y1}) to provide a systematic overview all squares, patches, holes, and micrographs within a defined region of a cryo-EM grid. Specifically, aldolase at a concentration of 1.6 mg/ml was dispensed on a support grid and prepared using a Vitrobot. Instead of picking the most promising squares and holes, we randomly selected 31 squares across the whole grid and imaged almost all the holes in these selected squares. This resulted in a dataset of 4017 micrographs from holes in these 31 squares. Overall, the data quality was poor, given that only 33.4\% of the micrographs have a CTF below 6 \AA. However, this makes the dataset very suitable for developing and testing algorithms for data collection algorithms, because 1) a perfect algorithm will aim to find the best data from mostly bad micrographs, and 2) the "unbiasedness" of this dataset ensures that when an algorithm selects a hole, the corresponding micrograph, and its metric can be provided as feedback.

In addition, we collected another different dataset (\textbf{Y2}) of 3969 micrographs with a different sample and grid type. We split both datasets into training and validation sets by a ratio of 2:1. In the experiments below, we evaluate our approach mainly based on Y1 while using Y2 to test the transferribility of CryoRL.

\noindent\textbf{Training and Evaluation}
We used the Tianshou reinforcement learning framework~\cite{weng2021tianshou} to learn \cryoRL. Each model was trained with 20 epochs, using the Adam optimizer and an initial learning rate of 0.01. We set the duration in our system to 240 minutes for training, and evaluate the system at 120, 240, 360 and 480 minutes, respectively.

\subsection{Main Results} 
\label{subsec:main_results}

\textbf{Comparison with Baselines.} We first developed a greedy-based method purely based on the hole classification results. This method performs a primary sorting on the grid-level images by their quality (i.e., the total number of low CTF holes), followed by a secondary sorting on the patches within a grid by the quality of patches. The sorted patches are then scanned in order, with only the holes classified as low CTFs visited. While simple, this greedy approach serves as a strong baseline when the hole-level classifier is strong.

We also compare our approach with two other widely used optimization techniques in practice: Genetic Algorithms (GA)~\cite{weise2009global} and Simulated Annealing (SA)~\cite{kirkpatrick1984optimization}. In these two solvers, solutions are sampled at the patch level rather than at the hole level for efficiency, and the fitness of the solutions are assessed according to the objective function proposed in this paper, i.e Eq.~\ref{eq:objective}. Since GA and SA are largely based on heuristic, the best solutions determined by them are scanned in a similar way to the greedy-based approach described above during the evaluation.

\begin{table}[htp]
    \centering
    \scriptsize
    \begin{adjustbox}{max width=\linewidth}
    \begin{tabular}{l|cccc}
        \toprule
       Methods&$\tau$=120&$\tau$=240&$\tau$=360&$\tau$=480\\
    \midrule
Random & 2.6$\pm$1.4& 5.1$\pm$1.6 & 7.3$\pm$2.3& 9.8$\pm$2.2\\
Greedy & 41.8$\pm$2.5& 69.3$\pm$3.2& 104.9$\pm$4.9& 147.9$\pm$5.1\\
Genetic Alg. (GA) ~\cite{weise2009global} & 28.3$\pm$6.5 & 72.3$\pm$6.8 & 115.7$\pm$7.8       &150.4$\pm$6.8 \\
Simulated Annealing (SA)~\cite{kirkpatrick1984optimization} & 39.4$\pm$6.5 & 73.3$\pm$7.0 & 104.7$\pm$8.9       & 147.9$\pm$9.6 \\
offline path planing~\cite{Sutton1998}& 44.3$\pm$0.9 &84.6$\pm$6.1 & 121.4$\pm$6.7       &166.6$\pm$4.9 \\
\midrule
CryoRL-DQN (ours) &41.7$\pm$3.1&86.6$\pm$3.0&\textbf{132.0}$\pm$2.3&171.4$\pm$2.0 \\
CryoRL-DQN$^{\dagger}$ (ours) &\textbf{47.4}$\pm$0.5&\textbf{89.0}$\pm$3.1&131.8$\pm$1.8&\textbf{172.6}$\pm$2.0 \\
\midrule
human  &31.9$\pm$10.6 &77.4$\pm$6.2 &- &-\\
\bottomrule
    \end{tabular}
    \end{adjustbox}
    \caption{\small Perf. comparison of CryoRL with baseline approaches on Y1}
    \label{table:baseline-comparison}
\end{table}

Table~\ref{table:baseline-comparison} reports the total number of low-CTF holes (\textit{\#lCTF}) found by each approach. For fair comparison, all the results are averaged over $50$ trials starting from random picked holes. Here, ResNet50 is used as the offline classifier, which achieves an accuracy around 83$\%$ in low-CTF hole classification (see Table~\ref{table:generalization}). The results based on ResNet18 can be found in the appendix. As shown in the table, our approach (\cryoRL-DQN) is clearly superior to all the baseline methods, producing quite promising results. With action elimination, the fast version of CryoRL (\cryoRL-DQN$^{\dagger}$) improve the performance further.
Note that while offline path planning yields comparable performance to our method, it is prohibitively costly in computation. 

To further illustrate the advantage of our approach, we plot for each approach the percentage of low-CTF holes over the total number of holes visited by time in Fig.~\ref{fig:cryo-efficency}. Our approach demonstrates high efficacy in data collection, finding $\sim$95\% of the holes in good quality. As a comparison, the percentage of low-CTF holes in Y1 is 33.4\% and the classification accuracy of low CTFs is only 83.9\%.

\begin{minipage}[!bt]{\textwidth}
  \begin{minipage}[b]{0.48\textwidth}
    \centering
    \includegraphics[width=0.75\linewidth]{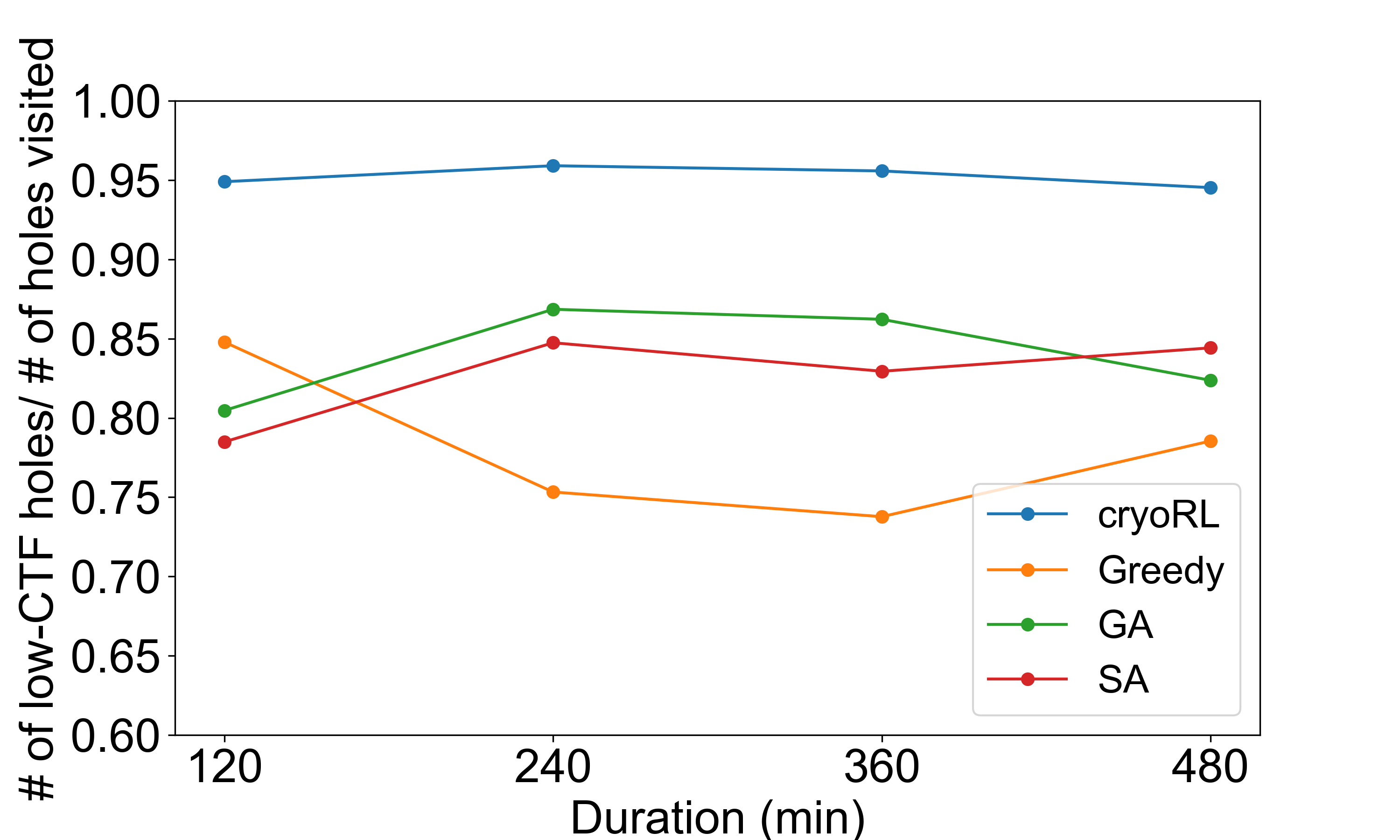}
    \captionof{figure}{Percentage of lCTF images visited during data collection. %
    }
    \label{fig:cryo-efficency} 
  \end{minipage}
  \hfill
  \begin{minipage}[b]{0.48\textwidth}
    \centering
    \includegraphics[width=0.75\linewidth]{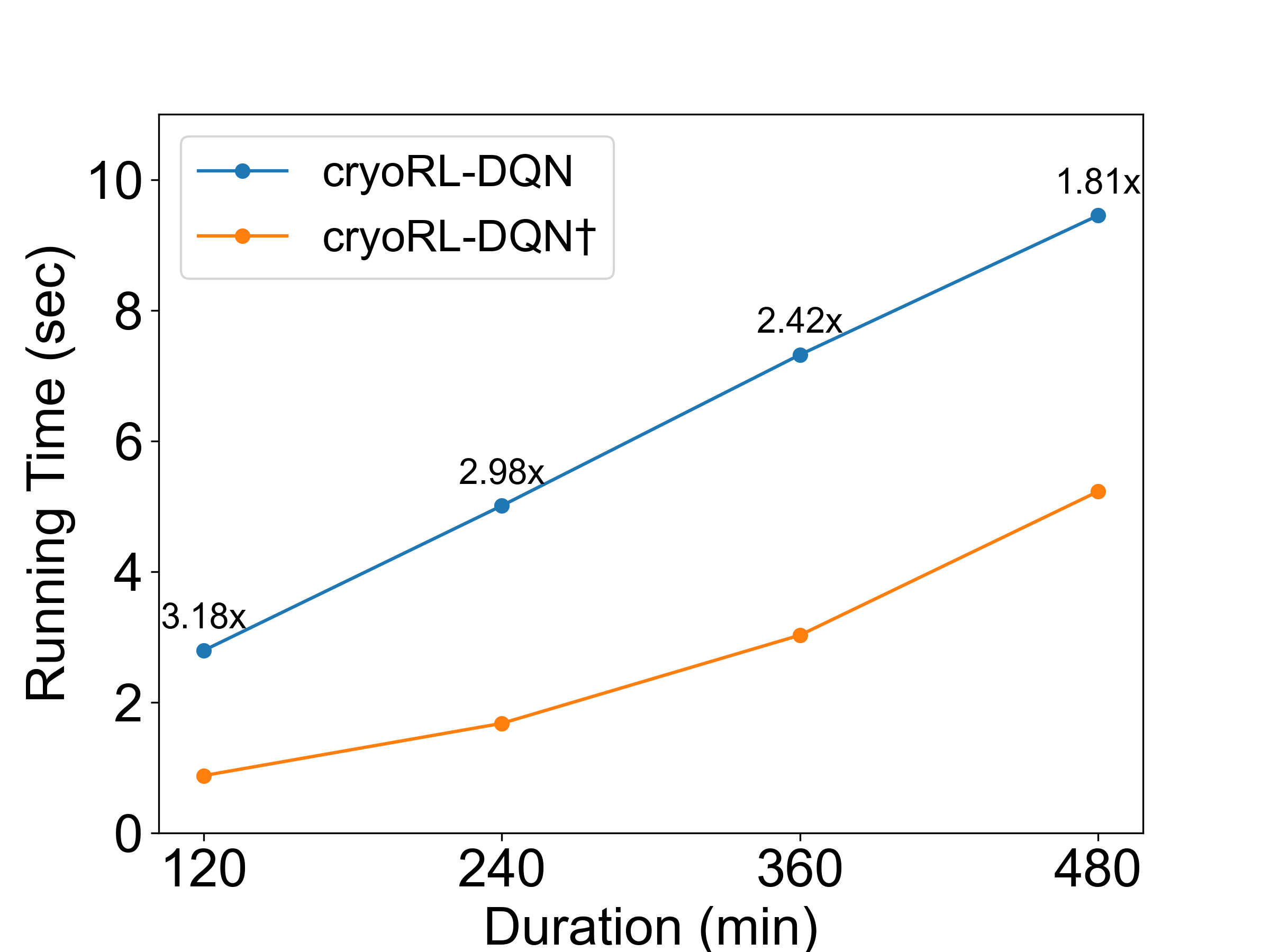}
    \captionof{figure}{Runtime comparison of CryoRL-DQN and its fast version with action elimination (CryoRL-DQN$^{\dagger}$). %
    } 
    \label{fig:cryo-runtime} 

    \end{minipage}
\end{minipage}

\setlength\intextsep{0pt}
\begin{table}[htp]
    \centering
    \scriptsize
    \begin{adjustbox}{max width=\linewidth}
    \begin{tabular}{l|llll}
        \toprule
       Methods&$\tau$=120&$\tau$=240&$\tau$=360&$\tau$=480\\
    \midrule
CryoRL-A2C &35.5$\pm$7.8&74.0$\pm$9.0&111.3$\pm$8.8&147.0$\pm$8.8 \\
CryoRL-C51 &39.4$\pm$4.2&76.3$\pm$3.1&109.6$\pm$2.0&141.0$\pm$2.7 \\
CryoRL-DQN &41.7$\pm$3.1&86.6$\pm$3.0&132.0$\pm$2.3&171.4$\pm$2.0 \\
CryoRL-DQN (dueling) &44.6$\pm$3.3&89.3$\pm$4.4&126.3$\pm$4.2&157.4$\pm$4.4\\
CryoRL-DQN (prioritized) &42.5$\pm$4.3&86.4$\pm$3.9&128.7$\pm$5.1&172.0$\pm$3.5\\
    \midrule
CryoRL-A2C$^{\dagger}$ &47.0$\pm$1.3\color{red}{(+32.3\%)}&90.8$\pm$4.0\color{red}{(+22.7\%)}&128.2$\pm$2.4\color{red}{(+15.1\%)}&163.9$\pm$4.7\color{red}{(+11.5\%)} \\
CryoRL-C51$^{\dagger}$ &47.4$\pm$0.9\color{red}{(+20.3\%)}&82.2$\pm$2.3\color{red}{(+7.7\%)}&116.9$\pm$1.0\color{red}{(+6.7\%)}&144.0$\pm$1.9\color{red}{(+2.1\%)} \\
CryoRL-DQN$^{\dagger}$ &\textbf{47.4}$\pm$0.5\color{red}{(+13.7\%)}&89.0$\pm$3.1\color{red}{(+2.8\%)}&131.8$\pm$1.8\color{red}{(+0.0\%)}&172.6$\pm$2.0\color{red}{(+1.0\%)} \\
CryoRL-DQN$^{\dagger}$ (dueling) &47.3$\pm$1.0\color{red}{(+6.1\%)}&89.4$\pm$2.9\color{red}{(+0.0\%)}&128.6$\pm$2.0\color{red}{(+1.8\%)}&165.4$\pm$2.5\color{red}{(+5.1\%)}\\
CryoRL-DQN$^{\dagger}$ (prioritized) &47.2$\pm$1.7\color{red}{(+11.1\%)}&\textbf{90.6}$\pm$2.8\color{red}{(+4.9\%)}&\textbf{132.9}$\pm$3.0\color{red}{(+3.3\%)}&\textbf{174.0}$\pm$3.1\color{red}{(+1.2\%)}\\
\bottomrule
    \end{tabular}
    \end{adjustbox}
    \caption{\small Perf. comparison of different CryoRL variants on Y1 ($\dagger$ indicates action elimination~(Section~\ref{subsec:dqn})). The performance gains from action elimination are highlighted by numbers in parentheses.}
    \label{table:more-comparison}
\end{table}

We also experimented with several
other RL variants including dueling DQN~\cite{wang2016dueling}, DQN with prioritized replay~\cite{schaul2015prioritized}, A2C~\cite{a2c} and C51~\cite{bellemare2017distributional}. 
As seen from Table~\ref{table:more-comparison}, the DQN family overall perform better than A2C and C51. Interestingly, A2C and C51 benefit substantially from action elimination and gain significant performance boosts, suggesting that restricting the actions to smaller valid sets helps these methods learn policies more effectively. CryoRL with action elimination also achieves considerable speedups in runtime by 2$\sim$3 times, as shown in Fig.~\ref{fig:cryo-runtime}. Since the performance differences between DQN models are minor, we focus on the vanilla DQN in the analysis below.

\textbf{Comparison with Human Performance}. We developed a simulation tool to benchmark human performance against the performance of \cryoRL. Fifteen students from two different cryo-EM labs with various expertise levels were recruited in this human study. The users did not have any prior knowledge of this specific dataset before participating in this study. Patch images containing holes in the same dataset were shown to the user. The user had either 50 or 100 chances to select the holes to take micrographs from, corresponding to the experiment's test duration of 120 or 240 minutes. After each selection, the CTF value for the selected hole was provided to the user. The goal of the users is to select as many "good" holes as possible in 50 or 100 chances. Note that we did not penalize the users for switching to a different patch or square as we did in \cryoRL. This encouraged the users to explore different patches initially and, theoretically, resulting in better performance than penalties applied. Nevertheless, we found that \cryoRL outperforms the human performance in both time durations  (Table~\ref{table:baseline-comparison}).

\setlength\intextsep{0pt}
\setlength\intextsep{0pt}
\begin{table}[tb]
    \centering
    \scriptsize
    \begin{adjustbox}{max width=\linewidth}
    \begin{tabular}{c|cc|ccc|cccc}
        \toprule
     
          \multirow{2}{*}{Test}    & \multicolumn{2}{c|}{Training}  & \multicolumn{3}{c|}{Top1 Acc.} & \multicolumn{4}{c}{\#lCTFs found}  \\
        \cmidrule{2-10}
                   &classifier & CryoRL & lCTF & hCTF   & all &$\tau$=120&$\tau$=240&$\tau$=360&$\tau$=480 \\
    \midrule
    Y1&Y1&Y1&83.9 &91.2 &88.5 &47.4$\pm$0.5&89.0$\pm$3.1&131.8$\pm$1.8&172.6$\pm$2.0 \\ 
    \midrule
    Y1&M&Y1&66.6 &85.1 &73.6 &44.7$\pm$2.2 & 70.0$\pm$4.7 &104.0$\pm$3.6 &138.9$\pm$2.6 \\
    Y2&M&Y2&69.5 &77.4 &73.5 &31.0$\pm$6.5 &56.3$\pm$7.7 &87.1$\pm$8.0 &125.5$\pm$8.3\\
    Y2&M&Y1&69.5 &77.4 &73.5 &20.9$\pm$4.3 &55.8$\pm$3.7 &83.1$\pm$3.5 &91.8$\pm$3.3 \\
    \bottomrule
    \end{tabular}
    \end{adjustbox}
    \vspace{-1mm}
    \caption{\small Generalization ability of the offline classifier and CryoRL. }
    \label{table:generalization}
\end{table}

\textbf{Transferrability.} 
 We further evaluate the \textit{transferability} of our proposed approach based on a new dataset \textbf{M},  which consists of users' daily use of microscope in real-life scenarios from 2019 to 2021. Different from Y1 and Y2 data where almost each hole in the patch images was imaged, M were only sparsely inspected, with a small portion of holes visited by the users. In other words, there are a lot of holes in the patch images without a CTF ground truth available. As a result, the limited coverage in M data is not sufficient for learning effective RL policies for planning microscope movement. Nevertheless, M data were collected under different realistic settings where various grid types and microscopes were used. It is much more diverse and substantially larger than Y1 data (over $100,000$ holes with CTF ground truth in M vs. ~$4,000$ in Y), making it suitable for building a foundation model for hole classification.

We split M data into training and validation sets at a ratio of 4:1 and trained a hole classifier based on Resnet50. We then applied the classifier to both test sets in Y1 and Y2, and the results are listed in Table~\ref{table:generalization}. As seen from the table, the classifier achieves moderate performance on Y1 and Y2, with an accuracy of around 70\% in low-CTF classification, suggesting that hole classification is still a challenging problem that needs further improvement.

We further trained RL models on Y1 and Y2 using the classification results based on the M model mentioned above.
As shown in Table~\ref{table:generalization}, a modest classifier (M) results in a performance drop in CryoRL ($4^{th}$ row) as expected, but the results are still reasonably good. Additionally, we extend to test the transferability of the RL models. Specifically, we applied the RL model based on Y1 to Y2 dataset and compared the results ($6^{th}$ row) to those from the RL model trained on Y2 itself ($5^{th}$ row). Even though Y1 and Y2 datasets were collected with different samples and grid types, the results between these two models are still comparable, showing the good transferability of CryoRL.

\subsection{Ablation Study}
\label{subsec:ablation}
In this section, we conduct experiments to characterize our proposed approach. We investigate how hole time duration and rewarding affects the performance of \cryoRL (i.e. the total number of low-CTF holes found ain a given amount of time). We also provide visualization of a planned path by CryoRL and the learned RL polices. %
  
\mycomment{

\textbf{Effects of classification accuracy}
\setlength\intextsep{0pt}
\begin{table}[tb]
    \centering
    \small
    \begin{adjustbox}{max width=\linewidth}
    \begin{tabular}{c|ccc|cccc}
        \toprule
       \multirow{2}{*}{Classifier} &  \multicolumn{3}{c|}{Top-1 Accuracy} & \multicolumn{4}{c}{low-CTF holes identified by \cryoRL} \\ 
    \cmidrule{2-8}
     &lCTF&hCTF&All&\tau=120 &\tau=240&\tau=360&\tau=480\\
    \midrule
R50^{*}  & 52.2&89&73.7&41.6 &77.6 &115.0 &144 \\
R50  & 83.9&91.2       &88.5&41.1 & 86.6 &132.0 & 171.4\\
    \bottomrule
    \end{tabular}
    \end{adjustbox}
    \vspace{-1mm}
    \caption{\small Effects of classifier accuracy on cryoRL performance (lCTF: low-CTF holes; hCTF:high-CTF holes)}
    \label{table:classification-accuracy-effect}
\end{table}
The hole-level classifiers based on Resnet18 and Resnet50 perform well on our data, achieving an accuracy of $\sim89\%$. To determine the effect of hole classification accuracy on cryoRL policy learning,  we trained a under-performing classifier R50$^{*}$ with a $\sim$73\% accuracy and applied it to learn \cryoRL. Table~\ref{table:classification-accuracy-effect} lists the top-1 accuracies of low-CTF and high-CTF holes based on different classifiers as well as the corresponding total number of lCTFs identified by \cryoRL under different time durations. As shown in the table, degraded performance in classification results in a performance drop in \cryoRL. Nevertheless, 
the comparable performance between \cryoRL-R50* and \cryoRL-R18 suggests that a modest classifier on low-CTF holes is sufficient for \cryoRL to converge on good holes as long as the classifier does not suffer from too many falsely classified low-CTF holes.

}

\setlength\intextsep{0pt}
\begin{table}[tb]
    \centering
    \caption{\textbf{Ablation study of CryoRL}}
    \label{table:jester_diving}
    \begin{subtable}[h]{0.48\textwidth}
            \centering
    \small
    \begin{adjustbox}{max width=\linewidth}
    \begin{tabular}{c|cccc}
        \toprule
       \multirow{2}{*}{Training} &  \multicolumn{4}{c}{Test Duration}  \\ 
    \cmidrule{2-5}
     Duration &$\tau$=120&$\tau$=240&$\tau$=360&$\tau$=480\\
    \midrule
    $\tau$=120 &40.4 & 82.1&123.1 &163.4 \\
    $\tau$=240 & 41.1&87.5&\textbf{130.0} & \textbf{165.5}\\
    $\tau$=360 &\textbf{45.7} &\textbf{90.2} &125.7 &163.5 \\
    \bottomrule
    \end{tabular}
    \end{adjustbox}
    \vspace{-1mm}
    \caption{\small Effects of time duration used in training on cryoRL performance.}
    \label{table:duration-effect}

    \end{subtable}
    \quad
    \begin{subtable}[h]{0.48\textwidth}
        
    \centering
    \small
    \begin{adjustbox}{max width=\linewidth}
    \begin{tabular}{c|c|cccc}
        \toprule
\multicolumn{2}{c|}{Rewards}&\multicolumn{4}{c}{Duration (minutes)} \\
    \cmidrule{1-6}
square-level &grid-level  & $\tau$=120 & $\tau$=240 & $\tau$=360 &$\tau$=480 \\ 
        \midrule
0.23 (default) &0.09 (default) &41.1&86.6 & \textbf{132.0} & 171.4  \\
0.23 ($\times 2$) &0.09  &\textbf{43.0}&\textbf{87.0} & 131.1 &\textbf{172.0} \\
0.23 &0.09 ($\times 2$) &41.6& 86.9 & 129.5 & 165.9  \\
0.23 ($\times 2$) &0.09 ($\times 2$) &41.8&80.8 & 124.7 & 163.3\\
    \bottomrule
    \end{tabular}
    \end{adjustbox}
    \vspace{-1mm}
    \caption{\small Effects of different rewards on \cryoRL's performance.}
    \label{table:reward-effect-8bit-new} 

    \end{subtable}
    \vspace{-3mm}
    \end{table}

\noindent \textbf{Effects of Time Duration}
In principle, the time duration $\tau$ used in training \cryoRL controls the degree of interaction of the RL agent with the data. A small $\tau$ limits \cryoRL to a few high-quality patches only, which might result in a more conservative policy that underfits. %
Table~\ref{table:duration-effect} confirms this potential issue, showing inferior performance when a short duration of $120$ minutes is used for training.

\mycomment{
\begin{table}[t]
    \centering
    \small
    \begin{adjustbox}{max width=\linewidth}
    \begin{tabular}{ccc|cccc}
        \toprule
\multicolumn{3}{c|}{Features}&\multicolumn{4}{c}{Duration (minutes)} \\
    \cmidrule{1-7}
training&test &score  & $\tau$=120 & $\tau$=240 & $\tau$=360 &$\tau$=480 \\ 
        \midrule
gt &pred &hard  &40.7 &85.8 & 123.5 &157.6 \\
pred &pred &hard &41.1&\textbf{87.5} & \textbf{130.0} & 165.5  \\
pred&pred &soft &\textbf{42.5} &81.5 & 127.2 & \textbf{165.9}\\
    \bottomrule
    \end{tabular}
    \end{adjustbox}
    \vspace{-1mm}
    \caption{\small Effects of different features on \cryoRL-R50's performance. \textit{gt}: ground truth; \textbf{pred}:prediction}
    \label{table:feaure-effect} 
\end{table}

\textbf{Performance of different features} The features we designed in Table~\ref{table:dqn-feaure} can be computed on either hard or soft hole-level categorization from the classifier. In addition, the training features can be based on hole-level categorization either from the true CTF values (\textit{gt}) or the classifier (\textit{pred}). We compare the performance of different feature combinations used for training and test in Table~\ref{table:feaure-effect}. From this analysis, we conclude that the model using hard categorization from the classifier for both training and test performs the best overall.
}

\noindent \textbf{Effects of Rewarding Strategies.} In our approach, the rewards used in policy learning are empirically determined. To check the potential impact of different rewards on the performance of \cryoRL, we trained more Q networks by doubling the reward for a) square switching; b) grid switching; and c) both. These changes are intended to encourage more active exploration of the data. As shown in Table~\ref{table:reward-effect-8bit-new}, the different rewarding schemes perform comparably, and increasing the reward for square switching leads to slightly better performance than the default setting. This suggests that CryoRL is not sensitive to rewards setup as long as the rewards value small microscope movement more. Nevertheless, how to optimize rewards for better performance of \cryoRL is an area of improvement in future work.

\setlength\intextsep{0pt}
\begin{figure}[!ptb]
    \centering
    \includegraphics[width=0.75\linewidth]{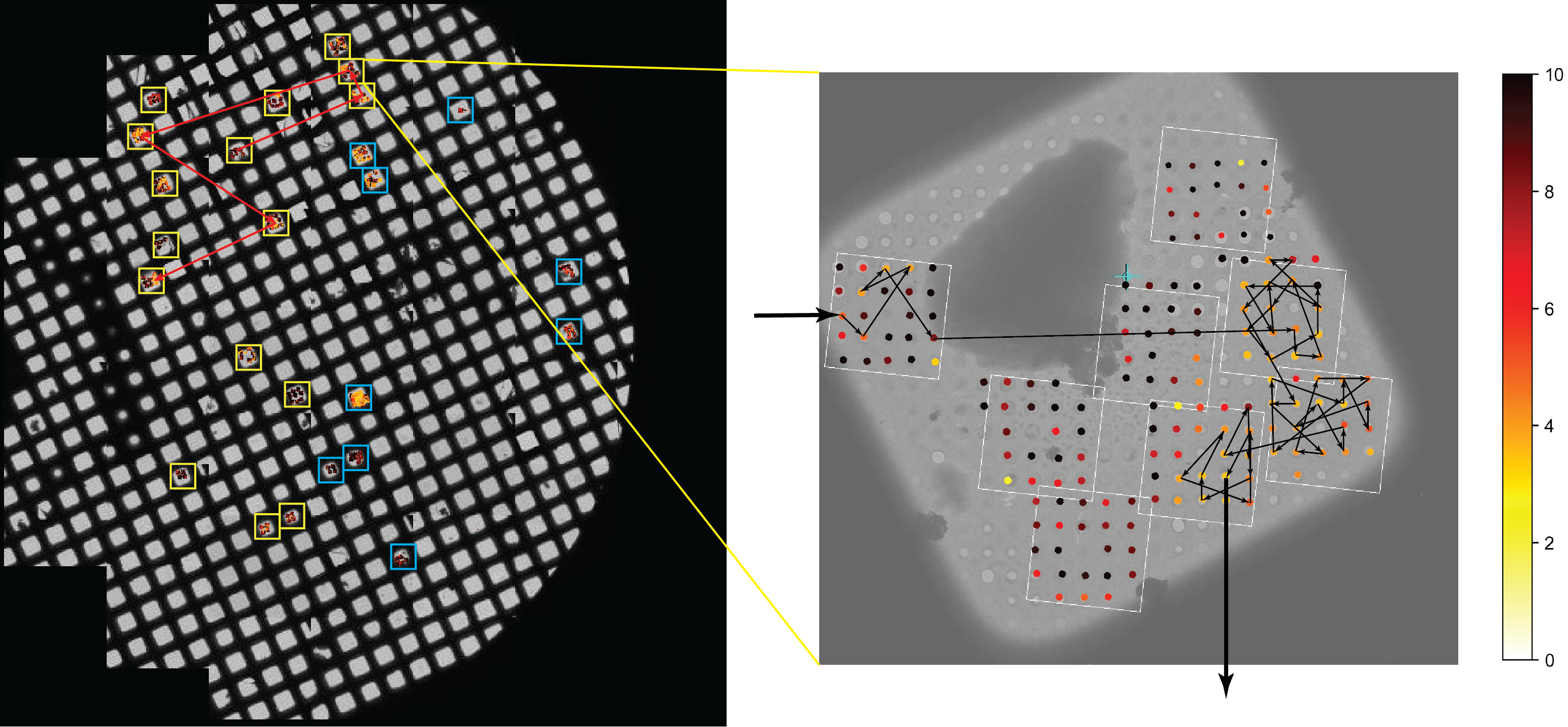}
    \caption{\small{A trajectory of microscope movement planned by CryoRL at square level (left) and patch level (right), respectively, in a 8-hour data collection session. The blue and yellow boxes show part of the training and validation sets while the color bar represents the ground-truth CTF value. The trajectory within a specific patch (right) illustrates that cryoRL can identify patches with more good holes (CTF$\le$6.0) in a global sense and prioritize their visits first. It is also noticed that some patches with a few good holes are left untouched in the square. This is because moving to a patch in another square (not shown here) is more rewarding than staying.}}
    \label{fig:two-level-trajectory} 
     \vspace{-3mm}
\end{figure}

\noindent \textbf{Trajectory Path and RL Policy Visualization.}
We plot one trajectory path of the microscope movement on the atlas planned by our CryoRL at square level (left) and patch level (right), respectively, in a 8-hour data collection session. The trajectory within a specific patch (right) illustrates that cryoRL can identify patches with more good holes (CTF$\le$6.0) in a global sense and prioritize their visits first. It is also noticed that some patches with a few good holes are left untouched in the square. This is because moving to a patch in another square (not shown here) is more rewarding than staying. 

\begin{figure}[!ptb]
    \centering
    \begin{tabular}{cc}
        \includegraphics[width=0.35\linewidth]{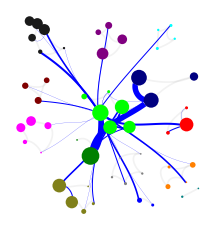} &
         \includegraphics[width=0.35\linewidth]{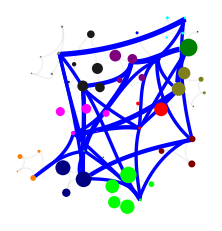} \\
          a) RL policy & b) user policy 
    \end{tabular}
    \caption{Illustration of data collection policies from cryoRL and human subjects. Here a graph node denotes a patch in our data and the size of the node indicates the quality of the patch (i.e the number of low-CTF holes). Patches from the same grid are grouped by color and linked by light grey edges. %
    A blue edge between a pair of patches shows how often the two patches are visited by the microscope. Intuitively, an effective policy should demonstrate strong connections between large-sized nodes, which is the case for the learned policy by our approach. As opposed to the RL policy, the human users presents random behaviors (b)).}
    \label{fig:RL-policy}
\end{figure}

We further compare and visualize the policies learned by our approach as well as the strategies used by human users. Specifically, we count how often the microscope visits a pair of hole-level images (i.e patches) in the 50 trials of our results and illustrate such information by an undirected graph. A node of the graph represents a patch and a blue edge between two patches indicates the frequency of them being visited by the microscope. Note that the node size here denotes the quality of a patch determined by the number of good holes in the patch, and the node color indicates the grid the patch belongs to. Intuitively, a good policy should show strong connections between large-sized nodes. As observed in  Fig~\ref{fig:RL-policy}a), our learned RL policy favors larger-size nodes, clearly demonstrating that CryoRL enables efficient data collection. Oppositely, the behavior of human users is random, with a lot of more patches being explored. This is because that the users were not penalized for switching different patches in the human study, and may also be due to the large variance in the user expertise.

\vspace{-3mm}
\section{Conclusion}
\vspace{-3mm}
\label{sec:conclusion}
To summarize, by combining supervised classification and deep RL, cryoRL provides a new framework for cryo-EM data collection. It can not only return the quality predictions for lower magnified hole level images but can also plan the trajectory for data acquisition. We have shown that \cryoRL combined with an offline hole classifier achieves better performance than average human users. Nevertheless, cryoRL needs squares to be pre-selected and all their corresponding patch-level images to be pre-captured. Future work will be needed to further optimize the RL system to consider more of this hierarchical process of cryo-EM data collection. The specific hyper-parameters, especially the penalties in the reward function, can also be improved for a more practical application. %

{\small
\bibliographystyle{plain}
\bibliography{reference,more_reference}
}

\newpage
\appendix
\section{Appendix}

\begin{table}[htp]
    \centering
    \scriptsize
    \begin{adjustbox}{max width=\linewidth}
    \begin{tabular}{l|l|llll}
        \toprule
       Methods&classifier&$\tau$=120&$\tau$=240&$\tau$=360&$\tau$=480\\
    \midrule
CryoRL-A2C &&37.0$\pm$6.7&71.9$\pm$9.5&104.1$\pm$9.4&144.8$\pm$9.3 \\
CryoRL-C51 &&37.2$\pm$4.2&70.6$\pm$5.0&98.1$\pm$5.0&128.0$\pm$3.6 \\
CryoRL-DQN &Resnet18&42.9$\pm$3.6&80.8$\pm$3.0&123.3$\pm$5.9&168.5$\pm$2.0 \\
CryoRL-DQN (dueling) &&42.9$\pm$4.2&86.9$\pm$5.2&125.2$\pm$5.3&159.5$\pm$6.9\\
CryoRL-DQN (prioritized) &&42.3$\pm$4.1&86.0$\pm$4.3&128.3$\pm$3.6&174.1$\pm$5.5\\
    \midrule
CryoRL-A2C$^{\dagger}$ &&46.0$\pm$2.6\color{red}{(+24.3\%)}&86.4$\pm$1.2\color{red}{(+20.2\%)}&124.4$\pm$2.2\color{red}{(+19.5\%)}&158.8$\pm$4.5\color{red}{(+9.7\%)} \\
CryoRL-C51$^{\dagger}$ &&46.5$\pm$0.8\color{red}{(+25.0\%)}&78.1$\pm$1.3\color{red}{(+10.6\%)}&116.7$\pm$1.0\color{red}{(+18.9\%)}&138.2$\pm$2.8\color{red}{(+8.0\%)} \\
CryoRL-DQN$^{\dagger}$ &Resnet18&\textbf{47.4}$\pm$2.0\color{red}{(+10.5\%)}&\textbf{91.0}$\pm$2.5\color{red}{(+12.6\%)}&132.8$\pm$2.1\color{red}{(+7.7\%)}&176.5$\pm$3.5\color{red}{(+4.7\%)} \\
CryoRL-DQN$^{\dagger}$ (dueling) &&47.2$\pm$1.0\color{red}{(+10.0\%)}&89.1$\pm$2.7\color{red}{(+10.0\%)}&129.2$\pm$1.8\color{red}{(+3.2\%)}&166.2$\pm$5.0\color{red}{(+4.2\%)}\\
CryoRL-DQN$^{\dagger}$ (prioritized) &&47.1$\pm$2.3\color{red}{(+11.3\%)}&90.4$\pm$2.3\color{red}{(+2.5\%)}&\textbf{133.0}$\pm$3.0\color{red}{(+3.7\%)}&\textbf{177.4}$\pm$4.1\color{red}{(+1.9\%)}\\
\midrule
CryoRL-DQN &&41.7$\pm$3.1&86.6$\pm$3.0&132.0$\pm$2.3&171.4$\pm$2.0 \\
CryoRL-DQN$^{\dagger}$ &Resnet50&47.4$\pm$0.5\color{blue}{(+13.7\%)}&89.0$\pm$3.1\color{blue}{(+5.1\%)}&131.8$\pm$1.8\color{blue}{(+0.0\%)}&172.6$\pm$2.0\color{blue}{(+1.0\%)}\\
\bottomrule
    \end{tabular}
    \end{adjustbox}
    \caption{\small Performance of different CryoRL variants on the Y1 dataset using Resnet18 as the offline hole classifier ($\dagger$ indicates action elimination.). The performance gains from action elimination are highlighted by numbers in parentheses. The numbers in bold mark the best performance achieved by CryoRL under different time durations using Resnet18 as the classifier.}
    \label{table:more-comparison-res18}
\end{table}

\noindent \textbf{Resnet18 Results.} We adopted Resnet18 as the offline classifier for CryoRL, which achieves better low-CTF classification accuracy than Resnet50 (91.0\% v.s 83.9\%), but lower high-CTF classification accuracy (87.5\% v.s 91.2\%). This suggests that Resnet18 yield more falsely classified good holes. As a result, CryoRL based on Resnet18 underperforms its counterpart based on Resnet50 (Table~\ref{table:more-comparison-res18}). However, when action elimination is applied, the performance of Resnet18 is significantly boosted and even gets slightly better than that of Resnet50. Additionally, action elimination greatly improves A2C and C51, similar to what's shown in the main paper.

\begin{table}%
\begin{subalgorithm}{.54\textwidth}
\begin{algorithmic}[1]
  \scriptsize
\Require States $S$, Actions $A$, Rewards $R$ 
\Require Learning Rate $\alpha$, Discounting factor $\gamma$, Elimination coefficient $beta$
\Require Switching costs $C$, Duration $\tau$
\Procedure {QLearning\_AE}{$S, A, R, C, \alpha, \beta, \gamma$, $\tau$}

\State $P \gets [p_0, p_1, \cdots, p_n]$ \Comment{Patches}
\State $L \gets [l_0, l_1, \cdots, l_n]$ \Comment{\# of predicted lCTFs in each patch}
\State $A' \gets Action\_Elim(P,L,C, \tau)$
\State $Q \gets QLearning(S, A', R, \alpha, \gamma)$ \Comment{standard Q\_learning}

\Return $Q$
\EndProcedure
\\
\\
\end{algorithmic}

\end{subalgorithm}%
\begin{subalgorithm}{.48\textwidth}
\begin{algorithmic}[1]
  \scriptsize
  \Procedure {Action\_Elim}{$P, L, C,\beta, \tau$}
\State $N_{max} \gets \beta * max\_lCTF($P$, $C$, \tau)$ \Comment{maximum lCTFs found assuming that all holes are good}
\State $n \gets 0$
\State $A' \gets \{\}$
\For{$p_i$ in $P$}
    \State $n \gets n + l_i$
    \State $A' \gets A'\bigcup \{h_j\in p_i| j=1 \cdots m_i\}$
    \If {$n \ge N_{max}$}
        \State $break$
    \EndIf
\EndFor

\Return $A'$
\EndProcedure
\end{algorithmic}
\end{subalgorithm}
\captionsetup{labelformat=alglabel}
\caption{Fast CryoRL with Action Elimination}
\label{alg:fast-CryoRL}
\end{table}

\noindent \textbf{Algorithm for Action Elimination.} The psudo code for action elimination is illustrated in Alg.~\ref{alg:fast-CryoRL}.
In the algorithm, \textbf{Action\_Elim} returns a list of valid actions, which are provided to the standard \textbf{QLearning} procedure or other policy learners for policy learning. The procedure \textbf{max\_lCTF} finds an upper limit of the number of low-CTF holes within a time duration $\tau$ under the assumption that all holes are in good quality. The elimination coefficient $\beta$ controls the size  of the valid action set. During training, $\beta$ should be set large to ensure sufficient training data with diversity. However, in test, $\beta$ can be set smaller to eliminate bad microscope movements while making action execution efficient.

\noindent \textbf{Experimental Setup for Genetic Algorithm (GA) and Simulated Annealing (SA)}
As mentioned in the main paper~(Section 5.2), the solutions of both GA and SA are assessed based on the same objective function used for RL, i.e Eq. 1 in the main paper. We implemented CryoRL-GA based on pyGAD~\cite{gad2021pygad} and Cryo-SA base on SimAnneal~\cite{simuatedanneal}. For CryoRL-GA, we set the number of generations to 40 and the solutions per population to 10. We use single-point crossover and and random mutation. For CryoRL-SA, the minimum and maximum temperatures are chosen as $1e-8$ and $\sqrt {N}$, respectively, where $N$ is the total number of training samples. The temperature reduction rate is set to 0.995.

\end{document}